Ruiqi Li[1], Liesbeth Allein[1], Damien Sileo & Marie-Francine Moens

# 7 When Do Discourse Markers Affect Computational Sentence Understanding?

**Abstract**: The capabilities and use cases of automatic natural language processing (NLP) have grown significantly over the last few years. While much work has been devoted to understanding how humans deal with discourse connectives, this phenomenon is understudied in computational systems. Therefore, it is important to put NLP models under the microscope and examine whether they can adequately comprehend, process, and reason within the complexity of natural language. In this chapter, we introduce the main mechanisms behind automatic sentence processing systems step by step and then focus on evaluating discourse connective processing. We assess nine popular systems in their ability to understand English discourse connectives and analyze how context and language understanding tasks affect their connective comprehension. The results show that NLP systems do not process all discourse connectives equally well and that the computational processing complexity of different connective kinds is *not always* consistently in line with the presumed complexity order found in human processing. In addition, while humans are more inclined to be influenced *during* the reading procedure but *not* necessarily in the final comprehension performance, discourse connectives have a significant impact on the final accuracy of NLP systems. The *richer* knowledge of connectives a system learns, the *more negative* effect inappropriate connectives have on it. This suggests that the correct explicitation of discourse connectives is important for computational natural language processing.

**Ruiqi Li**, KU Leuven & National University of Defense Technology, ruiqi.li@kuleuven.be

**Liesbeth Allein**, KU Leuven & European Commission, Joint Research Centre (JRC), liesbeth.allein@kuleuven.be

**Damien Sileo**, KU Leuven, damien.sileo@kuleuven.be

**Marie-Francine Moens**, KU Leuven, sien.moens@kuleuven.be

# 1 Introduction

Human language learning is a complex process involving many capabilities. Humans acquire vocabulary, order words in accepted sequences to form meaningful messages and consider the physical and social context to convey intentions and understand those of others. Computational systems — or models — working with natural language now aim to acquire such knowledge and proficiency with a single *pretraining* stage, during which they learn to perform a task requiring linguistic capabilities such as masked word prediction. Natural language processing (NLP) practitioners provide a randomly initialized model with a large number of texts in natural language and let it predict output for a predefined training task. For example, when the model is pretrained on the *masked word prediction task*, it tries to predict which word is hidden behind the mask in a given sentence: *The lioness hunts the [MASK] → zebra*. The correct answers are used to optimize the model and consequently improve its capabilities. By doing so, the model looks for patterns and important information in the input to adequately perform the pretraining task. These pretrained models can then be reused for other tasks. For example, NLP models can be used to tag the parts of speech in a sentence (Manning 2011; Plank, Søgaard, and Goldberg 2016), correct language-specific grammar errors (Heyman et al. 2018; Allein, Leeuwenberg, and Moens 2020), locate the correct answer to questions (Choi et al. 2018; Cartuyvels, Spinks, and Moens 2020) or even detect offensive or incorrect information (Ghadery and Moens 2020; Allein, Augenstein, and Moens 2021). To achieve high performance on these tasks, pretrained computational models need to acquire in-depth knowledge and understanding of syntax, semantics, pragmatics, and discourse. Given the scope of this volume, we focus on a special type of discourse markers, discourse connectives, and evaluate how they are leveraged by state-of-the-art NLP models to comprehend the discourse contained within a sentence. More specifically, we assess whether the presence or absence of discourse connectives affect a model's discourse understanding and compare the ease of comprehension between different connective types. Drawing parallels with the literature on human processing, this chapter also sheds light on the differences between humans and computers in terms of discourse comprehension. Last, we bridge the gap between computational and general linguistics by explaining and illustrating the core components in computational NLP models in a step-by-step manner.

---

[1] Joint first authorship.

The remainder of this chapter is structured as follows: Section 2 situates this chapter in the body of work that evaluated the discourse processing capabilities of humans and language processing abilities of computational NLP models; Section 3 illustrates step by step how these models turn language into a task output; Section 4 introduces the research questions and motivates them using findings from the linguistic literature on discourse connectives; Section 5 presents the nine sentence processing models and the three tasks we use in our evaluation; Section 6 elaborates on the methodology and contains the analyses and discussion; lastly, Section 7 concludes the chapter.

# 2 Related work

Our work contributes to the analysis of discourse markers in use. Discourse marker usage has been analyzed from the standpoint of corpus analysis (e.g., Schiffrin 2006; Prasad et al. 2008; Taboada 2006; Crible 2018). Many psycholinguistics studies focus on different methods by which discourse markers influence understanding. For example, Millis and Just (1994) and Haberlandt (1982) studied the influence of discourse markers on short-term word retention and reading speed, respectively. Degand, Lefèvre, and Bestgen (1999) and Degand and Sanders (2002) explored the influence of discourse relation marking understanding with question answering tests. Yung et al. (2017) developed a psycholinguistic model for discourse relation marking with an evaluation on the Penn Discourse Treebank dataset (Prasad et al. 2008). Studies on natural language processing models are valuable additions to psycholinguistics (Brysbaert, Keuleers, and Mandera 2014), as parallels can be established between NLP models and human processing (Mei et al. 2019; Ettinger 2020). Although many other evaluations have been proposed to assess their capabilities, our work is the first to assess the effect of discourse connectives on various computational models.

Sentence embedding models have been evaluated with many language understanding tasks, including natural language inference (NLI), sentence similarity, paraphrase detection, sentiment analysis, and question answering (e.g., Conneau and Kiela 2018; Wang et al. 2018, 2019). However, such results have been criticized for their lack of interpretability (Ribeiro et al. 2020). Conneau et al. (2018) proposed ten linguistic probing tasks: sentence length, word content, bigram shift, tree depth, top constituent, tense, subject number, object number, semantic odd man out, and coordination inversion. Wang et al. (2018) introduced a diagnostics-topic dataset that evaluates NLI accuracy when dealing with a more focused set of linguistic phenomena (lexical semantics, logic, knowledge, and propositional structure). Perone, Silveira, and Paula (2018) combined language understanding tasks and probing tasks and evaluated six sentence embedding models (ELMo, BoW, p-mean, Skip-Thoughts, InferSent and USE). They compared the embeddings on classification tasks (sentiment analysis, subjectivity/objectivity classification, question answering, and opinion polarity), semantic relatedness and textual similarity (image-caption retrieval, paraphrase detection, similarity and entailment), and the abovementioned ten linguistic probing tasks. Krasnowska-Kieras´ and Wróblewska (2019) evaluated six sentence embedding models (FastText, BERT, BiLSTM concatenated word and character embeddings, Sent2Vec, USE, and LASER) on some of the aforementioned linguistic probing tasks, other probing tasks (such as passive-active and sentence type classification), and two language understanding tasks (relatedness and entailment), and extended their evaluation to a multilingual setup. They compared sentence embedding performance for English and Polish. In contrast to the abovementioned research, we perform a more fine-grained evaluation of sentence embeddings by assessing their ability to grasp the meaning of connectives. Discourse connectives, including conjunctions and connective adverbials, have been leveraged for the training and evaluation of sentence embedding models (Jernite, Bowman, and Sontag 2017; Nie, Bennett, and Goodman 2019; Sileo et al. 2019, 2020) and there has been interest in measuring discourse coherence as a linguistic probing task when evaluating sentence embeddings (Chen, Chu, and Gimpel 2019).

However, these works focus on predicting a discourse connective or discourse relation between two sentences. Even though they show the value of discourse connective prediction as training/evaluation tasks, none of them measure the models' abilities to correctly make use of a connective within a sentence.

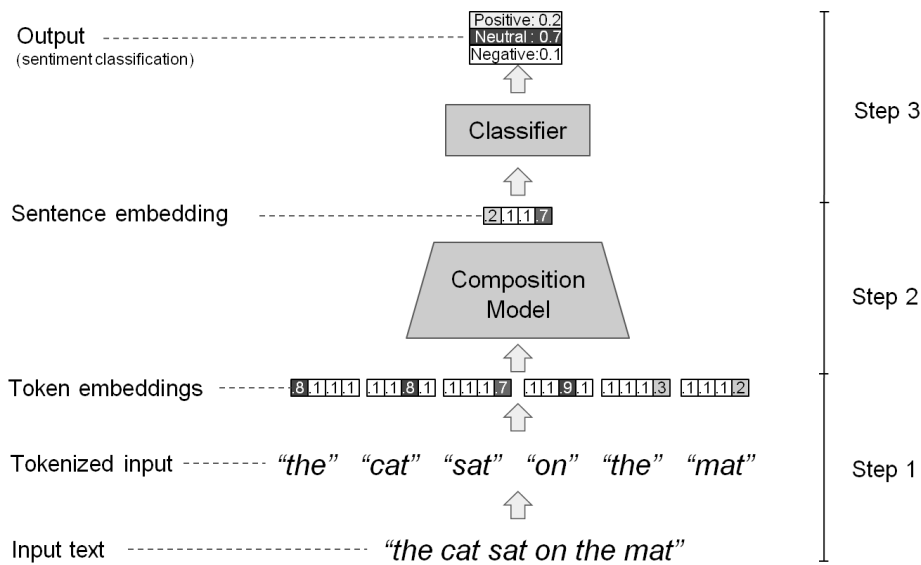

**Figure 1**: Overall architecture of a computational classification model.

# 3 Computational models: step-by-step

We first provide an overview of the three main processing steps involved in computational models performing sentence understanding tasks (0). Consider the two following sentences:

(1) *The lioness hunts the zebra.*

(2) *I go to the doctor because I am sick.*

## 3.1 From input to word representations

In order to process text, a computational model maps each word to a numerical word representation. Over the years, word representations have moved from categorical (Scott and Matwin 1999) to continuous representations (Naseem et al. 2020), illustrated in Figure 2. Categorical word representations are sparse[2], multidimensional vectors in which each dimension corresponds to a unique word in the vocabulary[3]. If we consider the two example sentences (1) and (2) as corpus, the vocabulary contains eleven words[4] and the categorical word vector for lioness is a vector with ten zero values and one non-zero value. The non-zero value can be either '1' (one-hot encoding), the word's absolute count in a sentence (Bag of Words), or its weighted importance compared with the occurrence frequency of other words in the entire corpus (term frequency-inverse document frequency, or TF-IDF). When adopting the TF-IDF approach, the non-zero value in the *lioness* word representation will be higher than the non-zero value in the word representation of *the* as the latter word is more frequent and therefore less unique. These categorical representations, however, do not encode a word's meaning a priori. For example, the categorical representations for *lioness* and *zebra* in Figure 2 do not reflect the relatedness we could expect between two mammals: the word *lioness* is equidistant[5] from both *zebra* and *sick*. Continuous word representations, or **word embeddings**, can encode such information. Instead of a specific word, the vector dimensions here represent word *aspects*. Continuous representations have a fixed dimensionality that is independent of the vocabulary size of a language. The word embeddings and their values are obtained by applying statistical learning techniques over a large collection of textual data. In a continuous multidimensional space, related words can be grouped, and unrelated words can be far apart. For example, when using the continuous representation, *lioness* is closer to *zebra* than to *sick* (Figure 2). The sentence representation models in this chapter represent words using

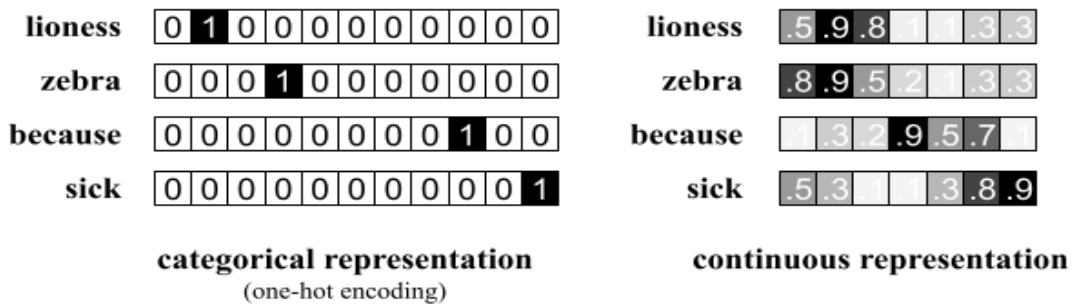

**Figure 2**: Examples of categorical and continuous word representations.

---

[2] Most dimensions of a vector are zeros, except for one dimension.
[3] NLP models usually have a fixed-size vocabulary that is extracted from a tokenized corpus and a minimum frequency criterion.
[4] the, lioness, hunts, zebra, I, go, to, doctor, because, am, sick
[5] Using metrics that measures the distance between two representations such as the Euclidean Distance.

continuous word embeddings.

## 3.2 From word representations to sentence representation

Once words are turned into numerical representations, a composition model now has to relate them to one another in a sentence in order to derive meaning. Given a sequence of word embeddings, a composition model computes a sentence representation, also called **sentence embedding**. Some sentence embedding models simply use the presence of words, and not their order or interaction to construct sentence representations. They often do this by taking the average of all the word embeddings in a sentence. Other models take into account the word order and relate the words to each other to derive the meaning of a sentence. For this, the models need to apply more complex statistical learning techniques. Sentence (2), illustrates the importance of processing word order. The causal connective (*because*) combines an action (*I go to the doctor*) and a situation (*I am sick*). The connective signals the motivation for the action. Both the action and situation refer to common real-world events, and the causal relationship between the two is plausible. Switching the position of the action and situation (i.e., *I am sick because I go to the doctor*) alters the original sentence's meaning.

State-of-the-art NLP systems use sentence embedding models to construct representations of natural language sentences. These models are trained using statistical learning techniques that incentivize the encoders to capture sentential semantics. In this chapter, we evaluate nine widely used sentence embedding models. More specifically, we look at how they behave on sentences with connective words.

## 3.3 From sentence representation to task prediction

After a sentence embedding model has constructed a sentence embedding, another computational model reasons over the sentence embedding to predict the desired outcome for a given task such as sentiment classification illustrated in Figure 1. This is generally done with a classifier using sentence embeddings as features. The classifier is an algorithm that implements a mathematical function to transform the multidimensional sentence embedding to the expected task output. When the model is learning a classification task, the classifier learns to relate features to the desired output labels based on annotated examples.

# 4 Research questions

We assess nine state-of-the-art sentence embedding models in their ability to represent sentences containing a discourse connective. We follow Fraser (1988) and regard discourse connectives as a category of discourse markers[6]. Discourse connectives signal a predicate between two discursive objects (Asher 1993), which are clauses in our work. We focus on discourse connectives because of their predominance in the datasets containing mainly written texts and the availability of semantic annotations (Prasad et al. 2008) of discourse connective roles. We address the following research questions:

1 Do sentence embedding models adequately capture the meaning of sentences containing a discourse connective?
2 How well do sentence embedding models deal with different connective types?
3 What is the effect of discourse connective removal on a sentence embedding?

---

[6] We do note that different acceptations of this term cohabit in the literature (Chen 2019).

4 If a connective is replaced with another connective, how much does this affect the sentence embedding?

**RQ 1** *Do sentence embedding models adequately capture the meaning of sentences containing a discourse connective?*

The ability to recognize and create connections between words and sentence parts to infer relations and meaning is crucial for text comprehension (Best et al. 2005; Crosson and Lesaux 2013). A person's mastery of discourse markers and reading comprehension accuracy is shown to be positively correlated (Khatib and Safari 2011). The presence of discourse connectives facilitates text comprehension, as they explicitly mark the relation between two sentence parts (Sanders and Noordman 2000; Degand and Sanders 2002). Nonetheless, humans still heavily rely on world and individual knowledge when inferring discourse meaning and relations (Singer and Gaskell 2007; Noordman and Vonk 2015), even when connectives are present (Noordman et al. 2015). Like humans, computational sentence embedding models need to have encountered various connectives denoting various relations when learning a language to recognize, understand and leverage connectives for meaning construction. The models also need to gain in-depth knowledge of the real world to verify whether the discourse relations are valid or to infer relations when discourse connectives are absent. The pretrained sentence embedding models discussed in this chapter learn that world from the millions of sentences they were exposed to in the pretraining phase.

We first compare the nine sentence embedding models' performances on three language understanding classification tasks for sentences containing a connective word. As we aim to directly evaluate the sentence representations, we always use the same simple classifier, namely, a logistic regression classifier. This ensures a controlled comparison of the sentence embedding models. We assume that the model's ability to adequately capture all the needed information contained within the sentences will translate into adequate performance on language understanding tasks. Therefore, low language understanding task performance indicates an inadequate encoding of connectives in the sentence embeddings. Conversely, high model performance indicates a sufficient integration of a connective's semantics.

**RQ 2** *How well do sentence embedding models deal with different connective types?*

Not all discourse relations are equally easy and fast to comprehend. The discourse connectives' relative cognitive complexities are often defined in terms of the order in which they are acquired in childhood. In this regard, Sanders (2005) hypothesized that causal relations are more complex than additive relations as additive connectives are first acquired when learning a language before causal connectives. That causal complexity hypothesis is supported by Bloom et al. (1980), who investigated the order in which English connectives appeared in the complex sentences of four children between two and three years old. They found that additive connectives are used earlier than those signaling temporal, causal, and adversative relations. Crosson, Lesaux, and Martiniello (2008) and Cain, Patson, and Andrews (2005) also confirmed that additive relations are cognitively the least complex of the four. Murray (1997) showed that the duration of human processing differs according to discourse connective types and that adversative connectives lead to the highest processing disruption. In contrast, Cain and Nash (2011) found that people process complex and specific connectives more quickly than the simpler, more general additive connective 'and', suggesting that higher complexity does not necessarily indicate longer processing or reading time.

We investigate deeper into the different types of connectives and compare sentence embedding behavior models in the presence of various connective types. We use the following connective types: additive, adversative, causal, and sequential[7] (Bernhardt 1980). Other categorizations of connectives or, more broadly, discourse markers have been adopted in the literature. For example, Fraser (1996) focused on the pragmatics behind discourse markers and categorized them as a type of pragmatic marker that only bears a procedural meaning and connects a message to a previous discourse. His proposed subcategories are topic change, contrastive, elaborative, and inferential markers. We refer to Alami (2015) for a more elaborate discussion on discourse marker categorizations. We hypothesize that sentence embedding models — like humans — will display differences in connective comprehension across the connective types.

**RQ 3** *What is the effect of discourse connective removal on a sentence embedding?*

Humans process explicit information more easily than implicit information (Florit, Roch, and Levorato 2011). When discourse connectives are absent and discourse relations are implicit, they need to rely on higher-order skills to fill in the gaps (Rapp et al. 2007). Research on the effect of discourse connectives on human processing polarizes. Some studies (Al-Surmi 2011; Cevasco, Muller, and Bermejo 2020) conclude that the presence of a discourse connective does not affect the accuracy of following reading comprehension tasks, but some other studies (Loman et al. 1983; Meyer, Brandt, and Bluth 1980) confirm that discourse connectives facilitate human text representation. Some researchers (Millis, Graesser, and Haberlandt 2010) even report a negative impact.

We remove connectives from the language understanding task input text and measure the influence of this removal on prediction accuracy. Drawing parallels with the negative effects of connective removal on human discourse understanding, we hypothesize that connective removal will negatively affect the sentence embedding models' discourse comprehension and task performance. If the accuracy drops, the sentence embedding models have learned to incorporate the meaning of the connectives when representing sentences. If it does not, this suggests that the sentence encoders ignore the connectives for sentence representation.

**RQ 4** *If a connective is replaced with another connective, how much does this affect the sentence embedding?*

Replacing a connective with another connective that signals a different discourse relation can lead to incoherence, thus making a sentence inexplicable or incomprehensible (Van Dijk 1977). Humans also need more time to process two-clause sentences when the connective signals an incorrect discourse relation that opposes the meaning of the two clauses (Cain and Nash 2011). As discussed in RQ1, both humans and computational systems need to rely on in-depth knowledge of the real world to verify whether a signaled discourse relation presents a state or event that can occur in the real world as perceived and acquired by the human or model.

We assess whether sentence embedding models can distinguish the discourse relations signaled by the connectives. We first replace each connective with a connective signaling a similar discourse relation. We expect that this substitution should not strongly affect model performance as the discourse relation remains fairly intact. We then substitute each connective with a connective signaling another discourse relation. We expect this switch to negatively affect model performance, as the sentence's meaning might be drastically altered and/or the sentence now presents an improbable reality.

---

[7] Instead of *sequential* connectives, the literature also uses temporal connectives; they can be considered synonymous as they both signal a sequence of events in time (Cain and Nash 2011).

# 5 Sentence embedding models and tasks

## 5.1 Sentence embedding models

**Table 1**: Sentence embedding models evaluated in the experiments.

| Class | Model |
|---|---|
| Word order invariant | BOW |
| | SIF |
| | p-Mean |
| Word order aware: RNN-based | SkipThought |
| | QuickThoughts |
| | DiscSE |
| | InferSent |
| Word order aware: Transformer-based | BERT |
| | USE |

Table 1 summarizes the nine sentence embedding models used in our study. We selected these particular models based on their popularity and their potential ability to capture the meaning of connectives. We group the sentence embedding models based on their encoding scheme: word order invariant (WO-invariant) and word order aware (WO-aware).

### 5.1.1 WO-invariant

WO-invariant embedding models represent a sentence by averaging all word embeddings (Figure 3). These models do not take into account the order of the input words.

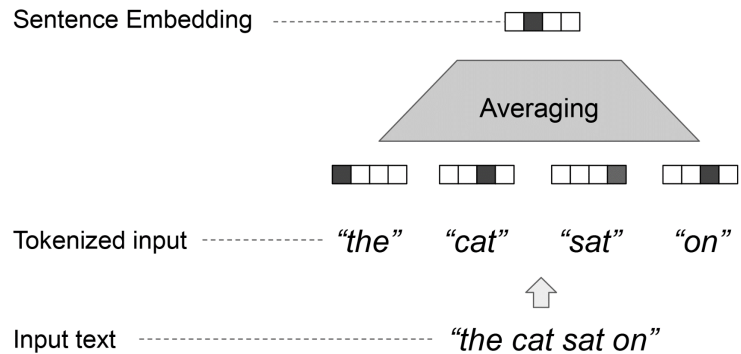

**Figure 3**: WO-invariant sentence embedding models: BOW, SIF, *p*-Mean.

**Continuous Bag Of Word (BOW)** is a widely-used baseline (Perone, Silveira and Paula 2018; Arora, Liang and Ma 2017; Tai, Socher and Manning 2015) which represents a sentence as the average of the embeddings of its words. In our experiments, we use two types of word embeddings: GloVe[8] and fastText[9], denoted as **BOW-GloVe** and **BOW-fastText**, respectively.

---

[8] The GloVe vectors have 300 dimensions and are trained on the 840 billion words Com-mon Crawl corpus, available at https://nlp.stanford.edu/projects/glove/
[9] The fastText vectors have 300 dimensions and are trained on the 600 billion words Com-mon Crawl corpus, available at https://fasttext.cc/docs/en/english-vectors.html

**Smooth Inverse Frequency (SIF)** model (Arora, Liang, and Ma 2017) is an extension of BOW where the word embeddings are weighted during averaging. The weight of a word is inversely proportional to the frequency of that word in the training corpus. Following Arora, Liang, and Ma (2017), we use GloVe as word embeddings (denoted as **SIF-GloVe**) and set $a$, the parameter that adjusts the influence of word frequency on word vector weight, to $a=0.0003$ as suggested by the authors.

***p*-power Mean Concatenation (p-mean)** technique (Rücklé et al. 2018) computes a sentence embedding with a geometric average of the word embeddings instead of the previously used arithmetic average. In our experiments, we use the four word embedding models provided by the authors and set of powers $p \in \{1, 2, 3\}$, denoted as ***p*-mean-1**, ***p*-mean-2**, and ***p*-mean-3**, respectively.

### 5.1.2 WO-aware

WO-aware sentence embedding models take into account the order in which word embeddings are presented (Figure 4).

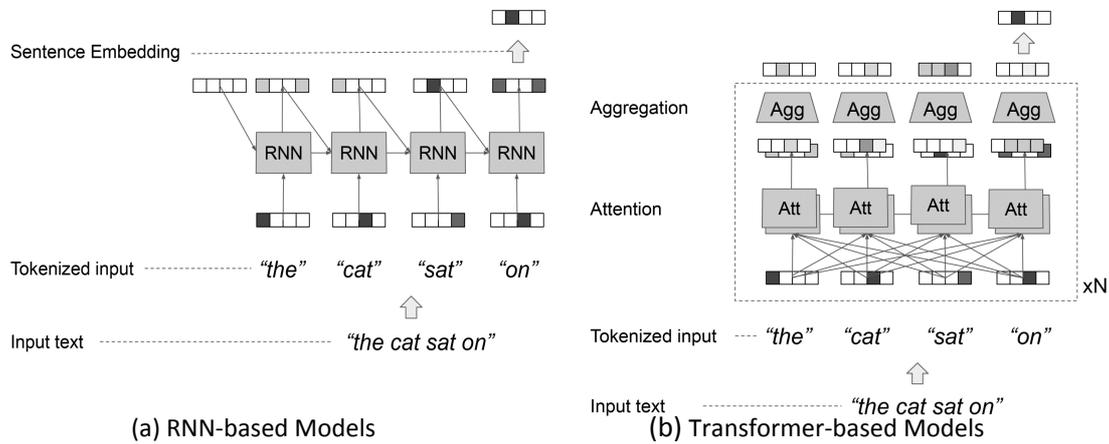

(a) RNN-based Models        (b) Transformer-based Models

**Figure 4**: WO-aware sentence embedding models. Their architecture can either be based on a recurrent neural network (RNN) or a Transformer.

**SkipThoughts** (Kiros et al. 2015) trains an encoder-decoder neural network to predict the previous and next sentences based on the current sentence. Both the encoder and decoder are Recurrent Neural Networks, namely Gated Recurrent Units (GRU). This model is trained on the BookCorpus (Zhu et al. 2015) dataset.

**QuickThoughts** (Logeswaran and Lee 2018) is similar to SkipThought, but here, the network has to predict the next sentence among a group of candidates from the further neighborhood. We use two QuickThoughts versions, one trained on the Book Corpus and the other on the UMBC webBase corpus (Han et al. 2013). We refer to them as **QuickThoughts-BC** and **QuickThoughts-UMBC**, respectively.

**DiscSE** (Sileo et al. 2019) is obtained with a RNN network that is trained to predict the discourse marker between two input sentences. The model is trained on a dataset derived from the Depcc corpus (Panchenko et al. 2018).

**InferSent** model (Conneau et al. 2017) builds sentence embeddings with a RNN network trained on a natural language inference task on the Stanford Natural Language Inference (SNLI) corpus (Bowman et al. 2015) and the Multi-Genre Natural Language Inference (MultiNLI) corpus (Williams, Nangia, and Bowman 2018). We use InferSent initialized with GloVe and fastText word embeddings as done in (Conneau et al. 2017). The resulting models are referred to as **InferSent-GloVe** and **InferSent-fastText**, respectively.

**BERT** (Devlin et al. 2019) is a bidirectional Transformer network trained with language modeling and next sentence prediction objectives, using data from the BookCorpus and the English Wikipedia. In our experiment, we use the BERT-Base and BERT-Large pretrained models containing 12 layers and 24 layers, respectively, and denoted as **BERT-12** and **BERT-24**.

**Universal Sentence Encoder** (USE) (Cer et al. 2018) is a Transformer encoder trained on multiple tasks including unsupervised tasks[10] (next sentence prediction and a conversational input-response task) and one supervised task (natural language inference). We refer to this model as **USE**.

## 5.2 Tasks

The sentence embedding models are evaluated using the three following language understanding tasks: sentiment classification, paraphrase detection, and coordination inversion detection. Each task evaluates a different aspect of sentence understanding. Table 2 presents an overview of the tasks and datasets used in our experiments.

**Table 2**: Overview of the tasks and datasets used in the extrinsic evaluation of sentence embeddings.

| Task | Dataset | #Test examples | #Examples with connective words | #Labels |
|---|---|---|---|---|
| Sentiment classification | MR | 10662 | 7635 | positive (5331) negative (5331) |
| | SST2 | 1821 | 1245 | positive (909) negative (912) |
| | CR | 3775 | 2603 | positive (2407) negative (1368) |
| | OSCAR | 2689 | 1970 | positive (923) negative (1320) mix(174) neutral (272) |
| Paraphrase detection | MRPC | 1725 | 1244 | paraphrase (1147) non-paraphrase (578) |
| | PAWS | 8000 | 5675 | paraphrase (3536) non-paraphrase (4464) |
| Coordination inversion detection | CoordInv | 10002 | 10002 | invalid (5001) valid (5001) |

**Sentiment classification** For a given sentence, the goal is to predict whether the sentence conveys a neutral, positive, or negative sentiment. The sentiment of a sentence depends on the words and the connections between them. Consider the following sentences:

(3) *The lioness hunts the zebra.* (neutral)
(4) *The magnificent and fearless lioness hunts the zebra.* (positive)
(5) *I like this movie because it contains violent scenes.* (positive-positive)
(6) *I like this movie although it contains violent scenes.* (positive-negative)

---

[10] The data is crawled from a variety of web sources including Wikipedia, Web news, Web question-answer pages and discussion forums.

Sentence (3) is neutral because it simply describes an action without presenting an opinion or a sentiment. By adding words that convey a positive sentiment (magnificent, fearless), the sentiment of the entire sentence becomes positive (sentence (4)). Similarly, discourse connectives can influence the sentiment of a sentence or clause. In sentence (5), violent scenes are considered a positive aspect of the movie — or even movies in general — and reinforce the overall positive sentiment of the entire sentence. On the contrary, this aspect is presented as negative by using an adversative instead of causal connective in sentence (6). As a consequence, the overall positive sentiment towards the movie is weakened.

**Coordination inversion detection**   The goal is to predict the natural ordering of two clauses. Given sentences composed of two coordinate clauses and a binding connective, 50% of the sentences are modified by inverting the position of the two clauses. The task is to predict whether the sentence was modified and thus *invalid*, or not (*valid*). Consider the following two sentences:

(7)  *I go to the doctor because I am sick.* (valid)
(8)  *I am sick because I go to the doctor.* (invalid)

Sentence (7) could naturally occur in a corpus, and sentence (8) is the coordination inversion of it. Due to the inversion, sentence (8) does not follow a plausible reasoning. Sentence embedding models should correctly encode semantic relations between coordinate clauses in order to fully solve this task. Discourse markers have a clear influence on this task, since replacing *because* by *therefore* in sentence (8) would make it much harder to detect a coordination inversion.

**Paraphrase detection**   The goal is to predict whether two sentences convey the same meaning.

(9) *The lioness hunts the zebra.* vs. *The zebra hunts the lioness.* (non-paraphrase)
(10)  *I go to the doctor because I am sick.* vs. *I go to the doctor even though I am sick.* (non-paraphrase)
(11)  *I go to the doctor because I am sick.* vs. *I am not feeling well so I am going to see my doctor.* (paraphrase)

Once again, this task is sensitive to word order, as illustrated in sentence (9). This task is by definition sensitive to semantic changes: sentences (10) and (11) show that changing a discourse marker is enough to change the expected outcome of a paraphrase detection.

# 6  Evaluation and discussion

## 6.1  Methodology

Many studies have explored the characteristics of human discourse processing, with a focus on discourse markers. For example, the presence and absence of markers elicit different reading behaviors in humans (Canestrelli, Mak, and Sanders 2013; Cain and Nash 2011), and different types of discourse markers also provoke varying reading responses (Murray 1997; Cain and Nash 2011). In order to investigate the mechanisms associated with the processing of discourse connectives by sentence embedding models, we set up four experiments corresponding to the four research questions described in Section 4. Unlike human language learners who have standardized language proficiency assessment systems such as TOEFL and IELTS, sentence embedding models do not have a standardized assessment system. It is therefore necessary to test the overall language processing ability of such models for sentences containing discourse connectives through designed experiments. We first set up a general evaluation to measure the models' ability to process and comprehend discourse connectives (General evaluation; RQ1). In controlled setups, we further study the influence of the presence of various connectives on model performance (Connective type evaluation; RQ2) and evaluate the language understanding tasks under specific connective perturbations (Connective omission/switch; RQ3/RQ4).

We select one-word connectives from the lexicons of Halliday and Hasan (1976). A manual annotation by two annotators was conducted to ensure that our evaluation focuses on the intended function. For each sentence, the connective words are selected and labeled with one category. The reliability of the annotations is ensured by a high inter-annotator agreement score of 98.82% (estimated from 150 randomly selected sentences with 254 candidate connective words). To base our findings on sufficient data, for each dataset, we only consider connective words that belong to the 5% most frequent words. We adopt the standard train/validation/test splits for each dataset, where a sentence that contains two or more connectives will appear in all relevant subtest sets. For the CoordInv dataset, we evaluate the split of the test set in valid and invalid sentences. We report the average of 5 runs per task and perform one-tailed t-test comparisons when applicable.

## 6.2 General evaluation

RQ1: *Do sentence embedding models adequately capture the meaning of sentences that contain a connective?*

This general evaluation answers RQ1 by measuring the model's overall ability to deal with discourse connectives. We select sentences containing one or more connectives in each dataset and evaluate the error rates (complementary to one of the accuracies) of the nine different sentence embedding models for these sentences.

| Category | Model | MR | CR | SST2 | OSCAR | MRPC | PAWS | CoordInv |
|---|---|---|---|---|---|---|---|---|
| Word Order Invariant | BOW-GloVe | 0.23 | 0.21 | 0.20 | 0.37 | 0.27 | 0.43 | 0.47 |
| | BOW-fastTex | 0.22 | 0.20 | 0.18 | 0.35 | 0.27 | 0.44 | 0.48 |
| | SIF-GloVe | 0.44 | 0.31 | 0.40 | 0.42 | 0.32 | 0.43 | 0.50 |
| | p-mean-1 | 0.22 | 0.20 | 0.17 | 0.36 | 0.27 | 0.43 | 0.46 |
| | p-mean-2 | 0.36 | 0.32 | 0.21 | 0.48 | 0.29 | 0.43 | 0.50 |
| | p-mean-3 | 0.27 | 0.26 | 0.21 | 0.43 | 0.27 | 0.43 | 0.48 |
| Word Order Aware | SkipThought | 0.24 | 0.20 | 0.18 | 0.35 | 0.27 | 0.36 | 0.32 |
| | QuickThoughts-BC | 0.20 | 0.17 | 0.15 | 0.51 | 0.24 | 0.32 | 0.30 |
| | QuickThoughts-UMBC | 0.18 | 0.15 | 0.13 | 0.51 | 0.24 | 0.30 | 0.31 |
| | BERT-12 | 0.19 | 0.14 | 0.14 | 0.32 | 0.28 | 0.42 | 0.31 |
| | BERT-24 | 0.16 | 0.12 | 0.11 | 0.30 | 0.31 | 0.42 | 0.29 |
| | DiscSE | 0.22 | 0.16 | 0.17 | 0.39 | 0.25 | 0.27 | 0.31 |
| | InferSent-GloVe | 0.20 | 0.15 | 0.17 | 0.36 | 0.25 | 0.26 | 0.33 |
| | InferSent-fastText | 0.26 | 0.25 | 0.18 | 0.42 | 0.26 | 0.26 | 0.37 |
| | USE | 0.20 | 0.14 | 0.14 | 0.30 | 0.28 | 0.41 | 0.41 |

**Figure 5**: Average absolute error rate for each sentence embedding model per task (General evaluation). For each dataset, color darkness is proportional to error rate.

Figure 5 confirms that WO-aware sentence embedding models (e.g., QuickThoughts, BERT, InferSent) outperform WO-invariant models (e.g., SIF, p-mean). On average, the former has a 24% lower error rate than the latter. The advantage of WO-aware models is particularly noticeable for the coordination inversion task (CoordInv). On this task, the error rates of the WO-invariant models approximate the accuracy of a random guess (i.e., 50% accuracy), which is substantially higher than those for the WO-aware sentence embedding models.

This can be explained by the fact that WO-invariant methods and WO-aware models dramatically differ in structure: WO-invariant methods simply take the (weighted) average of word embeddings. Each word is independently processed, and swapping the word order does not result in a change in the final sentence embedding. Therefore, they suffer from a very limited ability to capture semantic relationships. However, WO-aware models are designed with more complex neural network structures with multiple layers that can handle complex patterns. Through layer-by-layer interaction, words convey information to each other. As a result, WO-aware models capture more subtle semantics between words and are better at processing words like discourse connectives that specialize in expressing the interplay of words/discourses.

**Conclusion** WO-aware sentence embedding models outperform WO-invariant models ($p < 10^{-3}$) in presence of discourse connectives.

## 6.3 Connective type evaluation

RQ2: *How well do sentence embedding models deal with different connective types?*

We now focus on individual connectives and evaluate the performance of the sentence embedding models for each connective type. The connectives are divided into four types based on the denoted discourse relation: additive, adversative, causal, and sequential (Bernhardt 1980). Although these types can be further divided into more specific subtypes, we opted for these four generalized types because each type should have a statistically sufficient number of samples and because they are often used in other studies of discourse markers' impact on readability.

We start by evaluating the general trend of error rate change caused by the presence of a specific connective. For sentence embedding model and dataset , we compute the connective-specific error rate when classifying sentences containing the connective (Eq 1) and the overall error rate for all sentences in all the benchmark test set of dataset (Eq 2). Subsequently, we calculate the error rate difference between and (Eq 3). For brevity, we compute all the average error rate change (Eq 4) of all three () WO-invariant and all six () WO-aware sentence embedding models for each dataset D.

The results (Figure 6) for each individual connective show great differences in error rates across the different connective types, suggesting that not every discourse relation is equally easy to process. Sequential connectives decrease the error rate in five out of six datasets that contain its kind (Figure 6(a)-(f)). In the sentiment analysis task (Figure 6(a)-(d)), the error rate of adversative connectives is the highest ($p < 0.003$ for all datasets). Adversative connectives such as *however*, *although*, and *but* are a source of confusion, showing that sentence embeddings overall have difficulties capturing the meaning of adversative connectives. The causal connective *so* is consistently associated with high error rates. Regarding additional connectives, *or* is more difficult to handle than *also* and *and*. Overall, the adversative and causal connectives lead to significantly higher error rates than the additive and sequential connectives ($p < 10^{-4}$).

We observe different results for the paraphrasing task (Figure 6(e)-(f)). Additive connectives *also* and *and* slightly increase the error rate while adversative connectives largely lower the error rate. In the coordination inversion detection task (Figure 6(g)-(h)), connective *or* again correlates with a higher error rate than *and* and *but* ($p < 10^{-4}$). Overall, the impact of connectives depends on task type. The sentiment analysis task yields performance differences between connective types that are similar to their presumed complexity in the linguistic literature: higher performance with additive and sequential connectives and lower performance with causal and adversative connectives. This is in line with the complexity order of connectives established by Bloom et al. (1980): additive - sequential - causal - adversative (low to high complexity). However, in paraphrase detection and coordination inversion, the additional ones become the most difficult to process. This suggests that the effect of different types of connectives on computational language processing varies with tasks: paraphrase detection and coordination inversion are very sensitive to word order. Next, a person's mastery of discourse markers and reading comprehension accuracy is said to present a positive correlation (Khatib and Safari 2011). However, our study notes that the same phenomenon does not hold with sentence embedding models. Although WO-aware models have a richer knowledge of discourse connectives than WO-invariant models — as established in the general evaluation — both of them are similarly influenced by the different connective types. This suggests that sentence embeddings can use discourse connectives as superficial cues instead of using them to articulate discourse units. Concerning discourse processing speed, we cannot explore the difference between humans and computational models, as in most NLP models, sentences are all transformed into vectors of the same dimension and with the same computational cost.

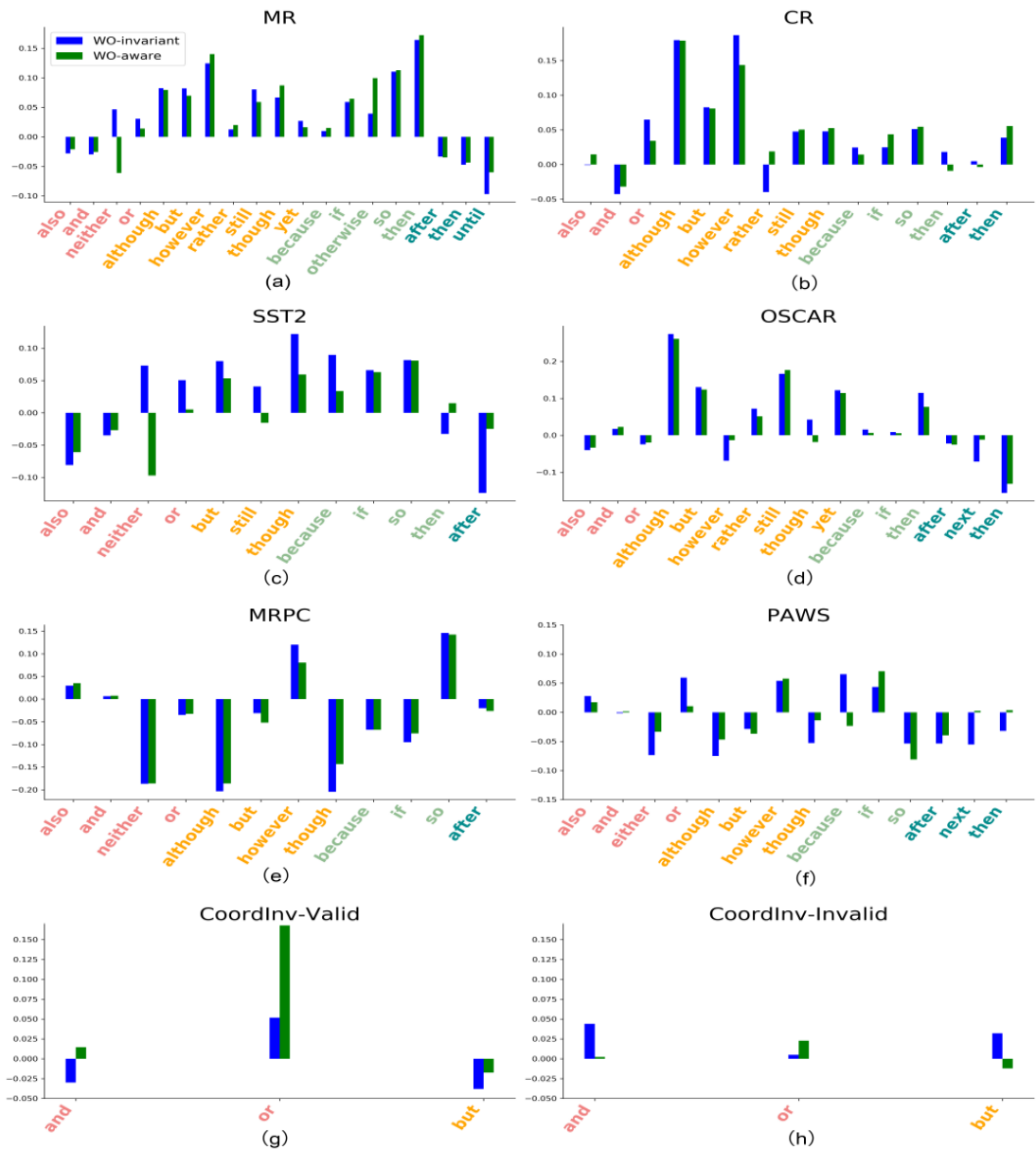

**Figure 5**: Error rate increase/decrease of WO-invariant (blue) and WO-aware (green) sentence embeddings for individual connectives (Connective type evaluation).

**Conclusion**   The sequential connectives are the easiest to handle across computation language understanding tasks. Adversative connectives are difficult to process in sentiment analysis tasks and additive ones are hard to deal with in paraphrase detection and coordination inversion.

## 6.4 Connective omission

RQ3: *What is the effect of discourse connective removal on sentence embeddings?*

We now modify connectives to directly evaluate their influence on sentence embeddings. For this, we leave out one connective in each input sentence by masking it with a special [MASK] token. This way, the sentence embedding models construct sentence embeddings that are not influenced by a discourse connective. We compare the error rate for all sentences in which the connectives are masked/omitted to the general error rate. The error rate change incurred by the omission is a measure of discourse connective importance. As connectives can strongly contribute to sentence meaning, we expect the error rate to increase in the case of connective omission.

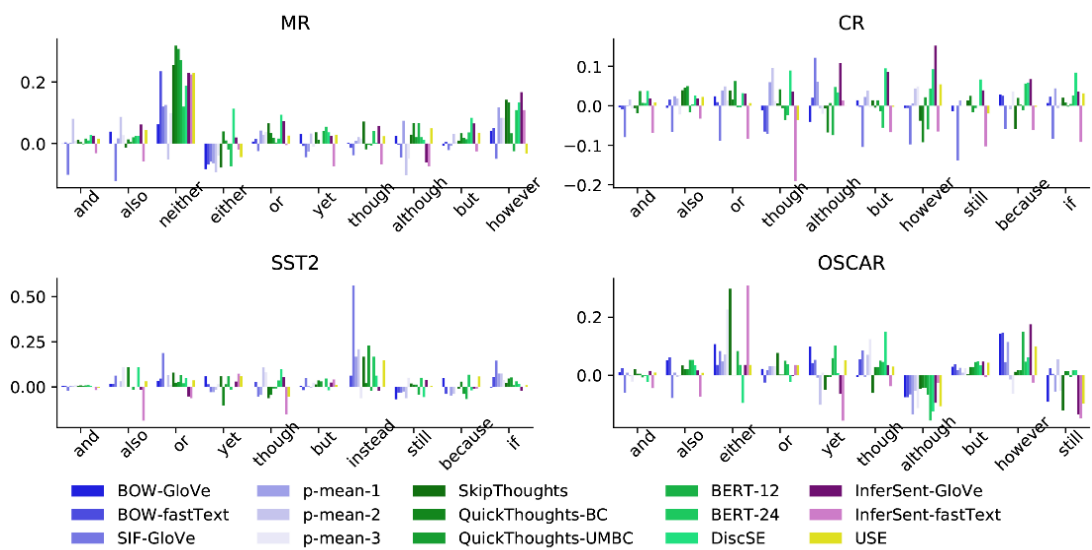

**Figure 5**: Error rate increase/decrease for each type of sentence embedding when corresponding connective words are removed (Connective omission).

Figure 7 displays the relative error rate variations when connectives are removed. Connective omission leads to an overall increase in error rates ($p < 10^{-6}$), suggesting that most sentence embedding models productively use the connective when computing embeddings. However, we observe that the effect of connective omission differs across the connective types. For example, omitting adversative connectives (e.g., *however*, *although*) affects error rates more strongly than omitting additive connectives (e.g., *and*, *also*) — with the highest error rate increase for *though* and *instead* ($p < 10^{-4}$). This suggests that sentence embedding models either heavily rely on adversative connectives or infer additive discourse relations better than adversative relations when a connective is missing. Intriguingly, we detect little difference in error rate change across connective types between WO-invariant sentence embedding models, which do not allow for non-additive interactions between words, and the more expressive WO-aware models. This could be attributed to a spurious correlation, which is confirmed upon dataset inspection. For instance, 58% of SST-2 reviews containing *instead* have a negative sentiment while only 20% are positive.

This highlights further differences between computational processing and human processing. Degand and Sanders (2002) suggested that discourse-insensitive tasks cannot present the significant effect of discourse connectives on human text processing. However, removing connectives in all datasets in our study results in consistent changes in error rates of computational models, suggesting that discourse connectives are a major factor in computational text processing, even for discourse-insensitive tasks. This conclusion is consistent with the observations in Section 6.2 that the explicitation of discourse connectives are essential for current sentence embedding models to infer pragmatic relations. By contrast, humans use discourse markers in a flexible way. For example, Cuenca (this volume) argues that even in academic text translation, where literal translations are predominant, omission of discourse markers is significant.

**Conclusion** The sentence embedding models appear to encode intra-sentential discourse relations relying on both the connective and the other words in the sentence. The processing of an adversative relation is the most dependent on the connective presence. Without correct connectives, the performance of sentence embedding models decreases across language understanding tasks. The dependence on correct discourse connectives for computational language understanding systems is larger than that for humans.

## 6.5 Connective switch

RQ4: *When a connective is replaced with another connective, how much does this affect the sentence embedding?*

We observed in the previous section that the presence of a connective word can have a strong influence on the error rate. In order to assess whether the models fully exploit connectives and not the mere presence of a connective word, we measure how a switch in connective type influences model performance. For this, we switch a connective with a connective that conveys a different discourse relation (e.g., *because → but*). We specifically focus on changing causal connective *because* with additive connective *also*, sequential connective *then* and adversative connective *but* in the sentiment classification and paraphrase detection datasets (Figure 8).

Although connective omission has led to similar performance variations for WO-invariant and WO-aware sentence embedding models (Section 3.4), connective switches cause higher error rate increases in WO-aware models. This suggests that WO-aware sentence embedding models make more non-trivial use of the inappropriate connectives than WO-invariant models. Overall, switching connectives cause an increase in error rate for all switches — with the highest error rate increase for *because → but*. As observed in Section 6.4, we see that sentence embedding models are again most sensitive to adversative connective *but*. This sensitivity indicates that the models not only strongly rely on *but* when computing embeddings, but also correctly embed the discourse relation it conveys.

The connective switching results suggest similarities between computational models and humans when faced with unsuitable discourse connectives. However, inappropriate discourse relations caused by a switch in connective type affect humans and computational models differently: humans are more inclined to produce changes in reading behavior (Canestrelli, Mak, and Sanders 2013) and reading time (Cain and Nash 2011; Murray 1997; Loureda et al, this volume), but not always in eventual reading comprehension accuracy (Al-Surmi 2011; Cevasco, Muller, and Bermejo 2020). Computational models have the opposite behavior: inappropriate discourse connectives greatly increase the error rate in reading tasks. Further, humans with a deeper knowledge of discourse connectives are more likely to reject the inappropriate connectives (Cain and Nash 2011), but computation systems with a richer knowledge of discourse connectives are more likely to be negatively affected.

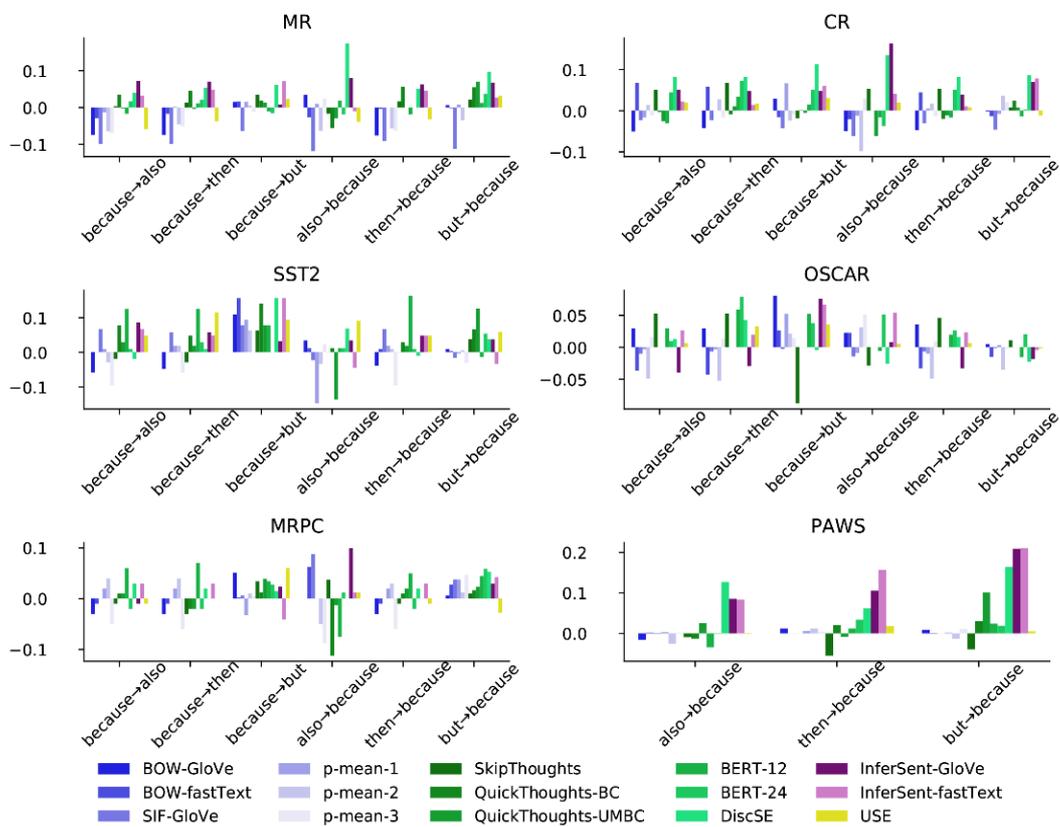

**Figure 5**: Error rate increase/decrease for each type of sentence embedding when corresponding connective words are switched (Connective switch).

**Conclusion** Substituting a connective with another connective that establishes a different discourse relation harms model performance, especially when it involves adversative connectives. The WO-aware sentence embedding models are more sensitive to connective switches than WO-invariant models. Contrary to humans, richer knowledge of discourse connectives makes a computational language understanding system more susceptible to the negative effects of incorrect connectives.

# 7 Conclusions

This chapter focused on assessing how connectives are leveraged in sentence embedding models and how the process differs from human processing.

As for the first general questions, our conclusions are: (a) The sequential connectives are the easiest to handle. Adversative connectives are difficult for computational sentiment analysis and addictive ones for order-sensitive tasks. (b) Omitting connectives actually influences computational language processing. Models with rich (WO-aware) and poor (WO-invariant) discourse connective knowledge are influenced to a similar extent, showing the connectives alone have a substantial predictive power. (c) The connective switch generally has a negative influence on computational language understanding. Systems with rich discourse knowledge suffer the most from inappropriate discourse connectives.

Concerning the difference between the processing mechanism for discourse connectives of computational models and humans: discourse connectives promote reading continuity and reduce the time that humans need to process utterances but they do not necessarily have an effect on reading comprehension accuracy, while the opposite is true for sentence embedding models. Current sentence embedding models cannot infer intersentential relations without discourse connectives as accurately as humans. Consequently, the presence and appropriate position of discourse connectives significantly reduces the error rate of sentence embedding models in subsequent classification tasks. In summary, our evaluation sheds light on the extent to which sentence embedding models make use of connectives and can be used to pinpoint the strengths and weaknesses of current models.

# 8 Acknowledgement

This work was realized with the collaboration of the European Commission Joint Research Centre under the Collaborative Doctoral Partnership Agreement No 35332. The research was designed and the experiments were conducted and analyzed when Liesbeth Allein was at KU Leuven; further analyses were performed and the chapter was written when she was at the European Commission. The scientific output expressed does not imply a policy position of the European Commission. Neither the European Commission nor any person acting on behalf of the Commission is responsible for the use which might be made of this publication. This work is also part of the CALCULUS[11] project: Damien Sileo and Marie-Francine Moens are funded by the ERC Advanced Grant H2020-ERC-2017 ADG 788506. Ruiqi Li is funded by the State Scholarship Fund of China Scholarship Council (CSC).

# References

Alami, Manizheh. 2015. Pragmatic functions of discourse markers: A review of related literature. *International Journal on Studies in English Language and Literature* 3(3). 1–10.

Allein, Liesbeth, Isabelle Augenstein & Marie-Francine Moens. 2021. Time-aware evidence ranking for fact-checking. *Journal of Web Semantics*. 100663.

---

[11] https://calculus-project.eu/

Allein, Liesbeth, Artuur Leeuwenberg & Marie-Francine Moens. 2020. Automatically correcting Dutch pronouns "die" and "dat". *Computational Linguistics in the Netherlands Journal* 10. 19–36.

Arora, Sanjeev, Yingyu Liang & Tengyu Ma. 2017. A simple but tough-to-beat baseline for sentence embeddings. In *Proceedings of 5th International Conference on Learning Representations (ICLR)*.

Asher, Nicholas. 1993. *Reference to abstract objects in discourse*. Vol. 50 (Studies in Linguistics and Philosophy). Springer Science & Business Media.

Bengio, Yoshua, R'ejean Ducharme, Pascal Vincent & Christian Janvin. 2003. A neural probabilistic language model. *The Journal of Machine Learning Research* 3. 1137–1155.

Bernhardt, Stephen A. 1980. Review of cohesion in English M.A.K. Halliday and R. Hasan. *Style* 14(1). 47–50.

Best, Rachel M, Michael Rowe, Yasuhiro Ozuru & Danielle S McNamara. 2005. Deep-level comprehension of science texts: The role of the reader and the text. *Topics in Language Disorders* 25(1). 65–83.

Bloom, Lois, Margaret Lahey, Lois Hood, Karin Lifter & Kathleen Fiess. 1980. Complex sentences: Acquisition of syntactic connectives and the semantic relations they encode. *Journal of Child Language* 7(2). 235–261.

Bowman, Samuel R., Gabor Angeli, Christopher Potts & Christopher D. Manning. 2015. A large annotated corpus for learning natural language inference. In *Proceedings of the 2015 Conference on Empirical Methods in Natural Language Processing (EMNLP)*, 632–642.

Brysbaert, M., Emmanuel Keuleers & Pawel Mandera. 2014. A plea for more interactions between psycholinguistics and natural language processing research. *Computational Linguistics in the Netherlands Journal* 4. 209–222.

Cain, Kate & Hannah Nash. 2011. The influence of connectives on young readers' processing and comprehension of text. *Journal of Educational Psychology* 103(2). 429–441.

Cain, Kate, Nikole Patson & Leanne Andrews. 2005. Age- and ability-related differences in young readers' use of conjunctions. *Journal of Child Language* 32(4). 877–892.

Canestrelli, Anneloes R., Willem M. Mak & Ted J. M. Sanders. 2013. Causal connectives in discourse processing: How differences in subjectivity are reflected in eye movements. *Language and Cognitive Processes* 28(9). 1394– 1413.

Cartuyvels, Ruben, Graham Spinks & Marie Francine Moens. 2020. Autoregressive reasoning over chains of facts with transformers. In *Proceedings of the 28th International Conference on Computational Linguistics (COLING)*, 6916–6930.

Cer, Daniel, Yinfei Yang, Sheng-yi Kong, Nan Hua, Nicole Limtiaco, Rhomni St John, Noah Constant, Mario Guajardo-Cespedes, Steve Yuan, Chris Tar, Brian Strope & Ray Kurzweil. 2018. Universal sentence encoder for English. In *Proceedings of the 2018 Conference on Empirical Methods in Natural Language Processing: System Demonstrations*, 169–174.

Cevasco, Jazmin. 2009. The role of connectives in the comprehension of spontaneous spoken discourse. *The Spanish journal of psychology* 12(1). 56–65.

Cevasco, Jazmin, Felipe Muller & Frederico Bermejo. 2020. Comprehension of topic shifts by Argentinean college students: Role of discourse marker presence, causal connectivity and prior knowledge. *Current Psychology* 39(3). 1072–1085.

Chen, Jiahuang. 2019. What are discourse markers? In *2nd Symposium on Health and Education 2019 (SOHE 2019)*, vol. 268 (Advances in Social Science, Education and Humanities Research), 1–9. Atlantis Press.

Choi, Eunsol, He He, Mohit Iyyer, Mark Yatskar, Wen-tau Yih, Yejin Choi, Percy Liang & Luke Zettlemoyer. 2018. QuAC: Question answering in context. In *Proceedings of the 2018 Conference on Empirical Methods in Natural Language Processing (EMNLP)*, 2174–2184.

Conneau, Alexis & Douwe Kiela. 2018. SentEval: An evaluation toolkit for universal sentence representations. In *Proceedings of the Eleventh International Conference on Language Resources and Evaluation (LREC)*.

Conneau, Alexis, Douwe Kiela, Holger Schwenk, Loic Barrault & Antoine Bordes. 2017. Supervised learning of universal sentence representations from natural language inference data. In *Proceedings of the 2017 Conference on Empirical Methods in Natural Language Processing (EMNLP)*, 670–680.

Conneau, Alexis, German Kruszewski, Guillaume Lample, Loic Barrault & Marco Baroni. 2018. What you can cram into a single &!#* vector: Probing sentence embeddings for linguistic properties. In *Proceedings of the 56th Annual Meeting of the Association for Computational Linguistics (ACL)*, 2126–2136.

Crible, Ludivine. 2018. *Discourse markers and (dis)fluency: forms and functions across languages and registers*. Vol. 286 (Pragmatics and Beyond New Series). John Benjamins Publishing Company.

Crosson, Amy C & Nonie K Lesaux. 2013. Does knowledge of connectives play a unique role in the reading comprehension of English learners and English-only students? *Journal of Research in Reading* 36(3). 241–260.

Crosson, Amy C, Nonie K Lesaux & Maria Martiniello. 2008. Factors that influence comprehension of connectives among language minority children from Spanish-speaking backgrounds. *Applied Psycholinguistics* 29(4). 603–625.

Degand, Liesbeth, Nathalie Lefevre & Yves Bestgen. 1999. The impact of connectives and anaphoric expressions on expository discourse comprehension. *Document Design* 1. 39–51.

Degand, Liesbeth & Ted Sanders. 2002. The impact of relational markers on expository text comprehension in L1 and L2. *Reading and Writing* 15(7). 739–757.

Devlin, Jacob, Ming-Wei Chang, Kenton Lee & Kristina Toutanova. 2019. BERT: Pre-training of deep bidirectional transformers for language understanding. In *Proceedings of the 2019 Conference of the North American Chapter of the Association for Computational Linguistics: Human Language Technologies (NAACL-HLT)*, 4171–4186.

Dolan, Bill, Chris Quirk & Chris Brockett. 2004. Unsupervised construction of large paraphrase corpora: exploiting massively parallel news sources. In *Proceedings of the 20th International Conference on Computational Linguistics*.

Ettinger, Allyson. 2020. What BERT is not: Lessons from a new suite of psycholinguistic diagnostics for language models. *Transactions of the Association for Computational Linguistics* 8. 34–48.

Florit, Elena, Maja Roch & M Chiara Levorato. 2011. Listening text comprehension of explicit and implicit information in preschoolers: the role of verbal and inferential skills. *Discourse Processes* 48(2). 119–138.

Fraser, Bruce. 1988. Types of English discourse markers. *Acta Linguistica Hungarica* 38(1/4). 19–33.

Fraser, Bruce. 1996. Pragmatic markers. *Pragmatics* 6(2). 167–190.

Ghadery, Erfan & Marie Francine Moens. 2020. LIIR at *SemEval* -2020 task 12: A cross-lingual augmentation approach for multilingual offensive language identification. In *Proceedings of the Fourteenth Workshop on Semantic Evaluation (semEval)*, 2073–2079.

Haberlandt, Karl. 1982. Reader expectations in text comprehension. *Advances in Psychology* 9. 239–249.

Halliday, Michael Alexander Kirkwood & Ruqaiya Hasan. 2014. *Cohesion in English*. Routledge.

Han, Lushan, Abhay L Kashyap, Tim Finin, James Mayfield & Jonathan Weese. 2013. UMBC EBIQUITY-CORE: Semantic textual similarity systems. In *Second Joint Conference on Lexical and Computational Semantics (*SEM), volume 1: proceedings of the main conference and the shared task: semantic textual similarity*, 44–52.

Heyman, Geert, Ivan Vulic, Yannick Laevaert & Marie-Francine Moens. 2018. Automatic detection and correction of context-dependent dt-mistakes using neural networks. *Computational Linguistics in the Netherlands Journal* 8. 49–65.

Hu, Minqing & Bing Liu. 2004. Mining and summarizing customer reviews. In *Proceedings of the 10th ACM SIGKDD International Conference on Knowledge Discovery and Data Mining*, 168–177.

Jernite, Yacine, Samuel R. Bowman & David Sontag. 2017. Discourse-based objectives for fast unsupervised sentence representation learning. *arXiv preprint arXiv:1705.00557*.

Khatib, Mohamad. 2011. Comprehension of discourse markers and reading comprehension. *English Language Teaching* 4. 243–250.

Kiros, Ryan, Yukun Zhu, Ruslan Salakhutdinov, Richard S. Zemel, Raquel Urtasun, Antonio Torralba & Sanja Fidler. 2015. Skip-thought vectors. In *Advances in Neural Information Processing Systems (NeurIPS)*, 3294–3302.

Krasnowska-Kieras, Katarzyna & Alina Woblewska. 2019. Empirical linguistic study of sentence embeddings. In *Proceedings of the 57th Annual Meeting of the Association for Computational Linguistics (ACL)*, 5729–5739.

Lai, Alice & Julia Hockenmaier. 2014. Illinois-LH: A denotational and distributional approach to semantics. In *International Workshop on Semantic Evaluation*.

Logeswaran, Lajanugen & Honglak Lee. 2018. An efficient framework for learning sentence representations. In *Proceedings of the 6th International Conference on Learning Representations (ICLR)*.

Loman, Nancy, L. Mayer, Richard & E. 1983. Signaling techniques that increase the understandability of expository prose. *Journal of Educational Psychology* 75(3). 402–412.

Manning, Christopher D. 2011. Part-of-speech tagging from 97% to 100%: is it time for some linguistics? In A F Gelbukh (ed.), *Computational Linguistics and Intelligent Text Processing. (CICLing) 2011*, vol. 6608 (Lecture Notes in Computer Science), 171–189.

Mei, Ning, Usman Sheikh, Roberto Santana & David Soto. 2019. How the brain encodes meaning: Comparing word embedding and computer vision models to predict fMRI data during visual word recognition. In *2019 Conference on Cognitive Computational Neuroscience*, 1088.

Meyer, Bonnie J F, David M Brandt & George J Bluth. 1980. Use of top-level structure in text: key for reading comprehension of ninth-grade students. *Reading Research Quarterly* 16(1). 72–103.

Millis, K. K., A. C. Graesser & K. Haberlandt. 2010. The impact of connectives on the memory for expository texts. *Applied Cognitive Psychology* 7(4). 317–339.

Millis, Keith K & Marcel Adam Just. 1994. The influence of connectives on sentence comprehension. *Journal of Memory and Language* 33(1). 128–147.

Murray, John D. 1997. Connectives and narrative text: The role of continuity. *Memory & Cognition* 25(2). 227–236.

Naseem, Usman, Imran Razzak, Shah Khalid Khan & Mukesh Prasad. 2021. A comprehensive survey on word representation models: From classical to state-of-the-art word representation language models. *Transactions on Asian and Low-Resource Language Information Processing* 20(5). 1–35.

Nie, Allen, Erin Bennett & Noah Goodman. 2019. DisSent: Learning sentence representations from explicit discourse relations. In *Proceedings of the 57th Annual Meeting of the Association for Computational Linguistics (ACL)*, 4497–4510.

Noordman, Leo G. M. & Wietske Vonk. 2015. Inferences in discourse, psychology of. In J D Wright (ed.), *International Encyclopedia of the Social & Behavioral Sciences (2nd ed.)* Vol. 12, 37–44. Elsevier.

Noordman, Leo G.M., Wietske Vonk, R Cozijn & S L Frank. 2015. Causal inferences and world knowledge. In *Inferences During Reading*, 260–289. Cambridge University Press.

Panchenko, Alexander, Eugen Ruppert, Stefano Faralli, Simone Paolo Ponzetto & Chris Biemann. 2018. Building a web-scale dependency-parsed corpus from CommonCrawl. In *Proceedings of the Eleventh International Conference on Language Resources and Evaluation (LREC)*.

Pang, Bo & Lillian Lee. 2005. Seeing stars: Exploiting class relationships for sentiment categorization with respect to rating scales. In *Proceedings of the 43rd Annual Meeting of the Association for Computational Linguistics*, 115–124.

Perone, Christian S., Roberto Silveira & Thomas S. Paula. 2018. Evaluation of sentence embeddings in downstream and linguistic probing tasks. *arXiv preprint arXiv:1806.06259*.

Plank, Barbara, Anders Søgaard & Yoav Goldberg. 2016. Multilingual part-of-speech tagging with bidirectional long short-term memory models and auxiliary loss. In *Proceedings of the 54th Annual Meeting of the Association for Computational Linguistics (ACL)*, 412–418.

Prasad, Rashmi, Nikhil Dinesh, Alan Lee, Eleni Miltsakaki, Livio Robaldo, Aravind Joshi & Bonnie Webber. 2008. The Penn Discourse TreeBank 2.0. In *Proceedings of the Sixth International Conference on Language Resources and Evaluation (LREC)*.

Rapp, David N, Paul van den Broek, Kristen L McMaster, Panayiota Kendeou & Christine A Espin. 2007. Higher-order comprehension processes in struggling readers: A perspective for research and intervention. *Scientific Studies of Reading* 11(4). 289–312.

Ruckle, Andreas, Steffen Eger, Maxime Peyrard & Iryna Gurevych. 2018. Concatenated *p*-mean word embeddings as universal cross-lingual sentence. *CoRR* abs/1803.01400.

Sanders, Ted. 2005. Coherence, causality and cognitive complexity in discourse. In *Proceedings of the First International Symposium on the Exploration and Modelling of Meaning*, 105–114.

Sanders, Ted JM & Leo G.M. Noordman. 2000. The role of coherence relations and their linguistic markers in text processing. *Discourse Processes* 29(1). 37–60.

Schiffrin, Deborah. 2006. Discourse marker research and theory: revisiting 'and'. In K Fischer (ed.), *Approaches to Discourse Particles*, 315–338. Elsevier Oxford.

Sileo, Damien, Tim Van de Cruys, Camille Pradel & Philippe Muller. 2019. Mining discourse markers for unsupervised sentence representation learning. In *Proceedings of the 2019 Conference of the North American Chapter of the Association for Computational Linguistics: Human Language Technologies (NAACL-HLT)*, 3477–3486.

Sileo, Damien, Tim Van de Cruys, Camille Pradel & Philippe Muller. 2020. DiscSense: Automated semantic analysis of discourse markers. English. In *Proceedings of the 12th Language Resources and Evaluation Conference (LREC)*, 991–999.

Singer, Murray & MG Gaskell. 2007. Inference processing in discourse comprehension. In M G Gaskell (ed.), *The Oxford Handbook of Psycholinguistics*, 343–359. Oxford University Press Oxford, UK.

Socher, Richard, Alex Perelygin, Jean Wu, Jason Chuang, Christopher D. Manning, Andrew Y. Ng & Christopher Potts. 2013. Recursive deep models for semantic compositionality over a sentiment treebank. In *Proceedings of the 2013 Conference on Empirical Methods in Natural Language Processing (EMNLP)*, 1631–1642.

Al-Surmi, Mansoor. 2011. Discourse markers and reading comprehension: Is there an effect? *Theory and Practice in Language Studies* 1(12). 1673–1678.

Taboada, Maite. 2006. Discourse markers as signals (or not) of rhetorical relations. *Journal of Pragmatics* 38(4). 567–592.

Tackstrom, Oscar & Ryan T. McDonald. 2011. Discovering fine-grained sentiment with latent variable structured prediction models. In *Proceedings of Advances in Information Retrieval - 33rd European Conference on IR Research*, 368–374.

Tai, Kai Sheng, Richard Socher & Christopher D. Manning. 2015. Improved semantic representations from tree-structured long short-term memory networks. In *Proceedings of the 53rd Annual Meeting of the Association for Computational Linguistics and the 7th International Joint Conference on Natural Language Processing of the Asian Federation of Natural Language Processing (ACL-IJCNLP)*, 1556–1566.

Van Dijk, Teun A. 1977. Semantic macro-structures and knowledge frames in discourse comprehension. *Cognitive Processes in Comprehension* 332. 3–31.

Wang, Alex, Yada Pruksachatkun, Nikita Nangia, Amanpreet Singh, Julian Michael, Felix Hill, Omer Levy & Samuel R. Bowman. 2019. SuperGLUE: a stickier benchmark for general-purpose language understanding systems. In *Proceedings of the 33rd International Conference on Neural Information Processing Systems (NeurIPS)*, 3266–3280.

Wang, Alex, Amanpreet Singh, Julian Michael, Felix Hill, Omer Levy & Samuel Bowman. 2018. GLUE: a multi-task benchmark and analysis platform for natural language understanding. In *Proceedings of the 2018 EMNLP Workshop BlackboxNLP: Analyzing and Interpreting Neural Networks for NLP*, 353–355.

Williams, Adina, Nikita Nangia & Samuel Bowman. 2018. A broad-coverage challenge corpus for sentence understanding through inference. In *Proceedings of the 2018 Conference of* *the North American Chapter of the Association for Computational Linguistics: Human Language Technologies (NAACL-HLT)*, 1112–1122.

Yung, Frances, Kevin Duh, Taku Komura & Yuji Matsumoto. 2017. A psycholinguistic model for the marking of discourse relations. *Dialogue & Discourse* 8(1). 106–131.

Zhang, Yuan, Jason Baldridge & Luheng He. 2019. PAWS: Paraphrase adversaries from word scrambling. In *Proceedings of the 2019 Conference of the North American Chapter of the Association for Computational Linguistics: Human Language Technologies, NAACL-HLT 2019, Minneapolis, USA, June 2-7, 2019, Volume 1 (long and short papers)*, 1298–1308.

Zhu, Yukun, Ryan Kiros, Rich Zemel, Ruslan Salakhutdinov, Raquel Urtasun, Antonio Torralba & Sanja Fidler. 2015. Aligning books and movies: towards story-like visual explanations by watching movies and reading books. In *Proceedings of the IEEE International Conference on Computer Vision (ICCV)*, 19–27.